\documentclass{article}

\usepackage[preprint]{neurips_2019}

\usepackage[]{neurips_2019}

\usepackage[utf8]{inputenc} 
\usepackage[T1]{fontenc}    
\usepackage{hyperref}       
\usepackage{xcolor}
\hypersetup{      
                    colorlinks=true,                
                    breaklinks=true,                
                    urlcolor= black,                
                    linkcolor= black,                
                    bookmarksopen=false,
                    citecolor=blue,
                    filecolor=black,
                    linkbordercolor=blue
}
\usepackage{url}            
\usepackage{booktabs}       
\usepackage{amsfonts}       
\usepackage{nicefrac}       
\usepackage{microtype}      
\usepackage{graphicx}
\usepackage{amsmath}
\usepackage{amssymb}
\usepackage{amsthm}
\usepackage{graphicx}
\usepackage{wrapfig}
\usepackage{appendix}

\newtheorem{theorem}{Theorem}
\theoremstyle{remark}

\usepackage{color}

\title{Out-of-Distribution Detection Using Neural Rendering Generative Models}

\author{%
  Yujia Huang \thanks{Equal contribution.} \\
  California Institute of Technology\\
  \texttt{yjhuang@caltech.edu} \\
  \And
  Sihui Dai \footnotemark[1] \\
  California Institute of Technology \\
  \texttt{sdai@caltech.edu} \\
  \AND
  Tan Nguyen \\
  Rice University \\
  \texttt{mn15@rice.edu} \\
  \And
  Richard G. Baraniuk \\
  Rice University \\
  \texttt{richb@rice.edu} \\
  \And
  Anima Anandkumar \\
  California Institute of Technology \\
  \texttt{anima@caltech.edu} \\
}

\begin{document}

\maketitle

\begin{abstract}

Out-of-distribution (OoD) detection is a natural downstream task for deep generative models, due to their ability to learn the input probability distribution. There are mainly two classes of approaches for OoD detection using deep generative models, viz., based on likelihood measure and the reconstruction loss.
However, both approaches are unable to carry out OoD detection effectively,  especially  when the OoD samples have smaller variance than the training samples. 
For instance, both flow based  and VAE models assign higher likelihood to images from SVHN  when trained on CIFAR-10 images. 
We use a recently proposed generative model known as neural rendering model (NRM) and derive metrics for OoD. We show that NRM unifies both approaches since it provides a likelihood estimate and also carries out reconstruction in each layer of the neural network. Among various measures, we found  the joint likelihood of latent variables to be the most effective one for OoD detection. 
Our results show that when trained on CIFAR-10, lower likelihood (of latent variables) is assigned to SVHN images. Additionally, we show that this metric is consistent across other OoD datasets.
To the best of our knowledge, this is the first work to show consistently lower likelihood for OoD data with smaller variance with deep generative models.

\end{abstract}

\section{Introduction}
Out-of-distribution (OoD) detection is an important topic in  machine learning. In applications such as medical diagnosis, fraud detection, failure detection, and security, being able to detect OoD samples is crucial for safe and automated deployment of machine learning. In practice, it is often not known a priori what these abnormal distributions are, making OoD detection a difficult problem to solve. Since the models are often trained to give accurate predictions on a specific dataset, they are susceptible to making false predictions  on OoD data.   Deep neural networks tend to make false predictions with  high confidence, due to well known calibration issues~\cite{guo2017calibration}. This makes OoD a challenging task for deep learning.

Deep generative models aim to learn the underlying distribution of training data $p(x)$ and generate images that mimic the training data. These two aspects of deep generative models bring forth two possible metrics for OoD detection: likelihood and reconstruction loss. 

The underlying expectation of using reconstruction loss for OoD detection is that since generative models learn to generate images based on the in-distribution data, they should be able to reconstruct in-distribution data well.  Thus, reconstruction loss of in-distribution images should be smaller than that of OoD images. Reconstruction loss based OoD detection methods have been explored with auto-encoders ~\cite{zhou2017anomaly}, variational auto-encoders ~\cite{an2015variational} and generative adversarial networks (GAN) ~\cite{zenati2018efficient}~\cite{li2018anomaly}. A weakness of using reconstruction loss is that the underlying expectation does not always hold: due to the large capacity in deep generative models, they are able to reconstruct image sets with smaller variance, even if these images are OoD. Thus,  reconstruction loss is usually a poor metric for OoD detection.

Likelihood-based methods for OoD detection use the learned density $p_\theta (x)$ and hence, only apply to generative models that provide likelihood estimates.  In this sense, flow-based models are particularly attractive because of the tractability of exact log-likelihood for both input data and latent variables. A well calibrated model should assign higher likelihood for training data and lower likelihood for OoD data.  However, with deep generative models, this is not always the case. A previous study by ~\cite{nalisnick2018deep} showed that deep generative models such as Glow ~\cite{kingma2018glow} and variational auto-encoder (VAE) assign higher likelihood to SVHN when trained on CIFAR-10 (Figure~\ref{fig:top-25} (c, d)). 
This counter-intuitive behavior of the likelihood of deep generative models is a fundamental issue based on the second order analysis in \cite{nalisnick2018deep}. It turns out that the model would assign higher likelihood to OoD data if it has lower variance than training data in general, and there is no straightforward solution to this issue.

The weakness of both likelihood-based and reconstruction-based methods in the setting where OoD samples have lower variance than in-distribution samples calls for rethinking in the choice of deep generative models for OoD detection. 
We explore the use of neural rendering model (NRM), a recently proposed deep generative model \cite{DBLP:journals/corr/abs-1811-02657}.  It attempts to reverse the feed-forward process in a convolutional neural network (CNN). In NRM, uncertainty in the generation process is controlled by latent variables $z$ with a structured prior distribution $p(z|y)$ that captures the dependencies across the network layers. 
Unlike other generative models, NRM has an architecture that mirrors a CNN, making the latent variables potentially more informative about the image distribution. Also, we can employ state-of-art  CNN architectures such as the ResNet and DenseNet to obtain good performance.
In this paper, our contributions are as follows:
\paragraph{We unify reconstruction loss and likelihood based OoD methods into one framework.} The observed likelihood $p(x)$  in NRM can be decomposed into two terms: reconstruction loss and joint likelihood of latent variables. Similarly, the latent likelihood can be decomposed into a reconstruction loss and joint likelihood of latent variables in the layers above. Thus, we have a wide choice of metrics that 
can be used for OoD detection and other downstream tasks. 
\begin{figure} 
\centering
  \includegraphics[width=0.95\linewidth]{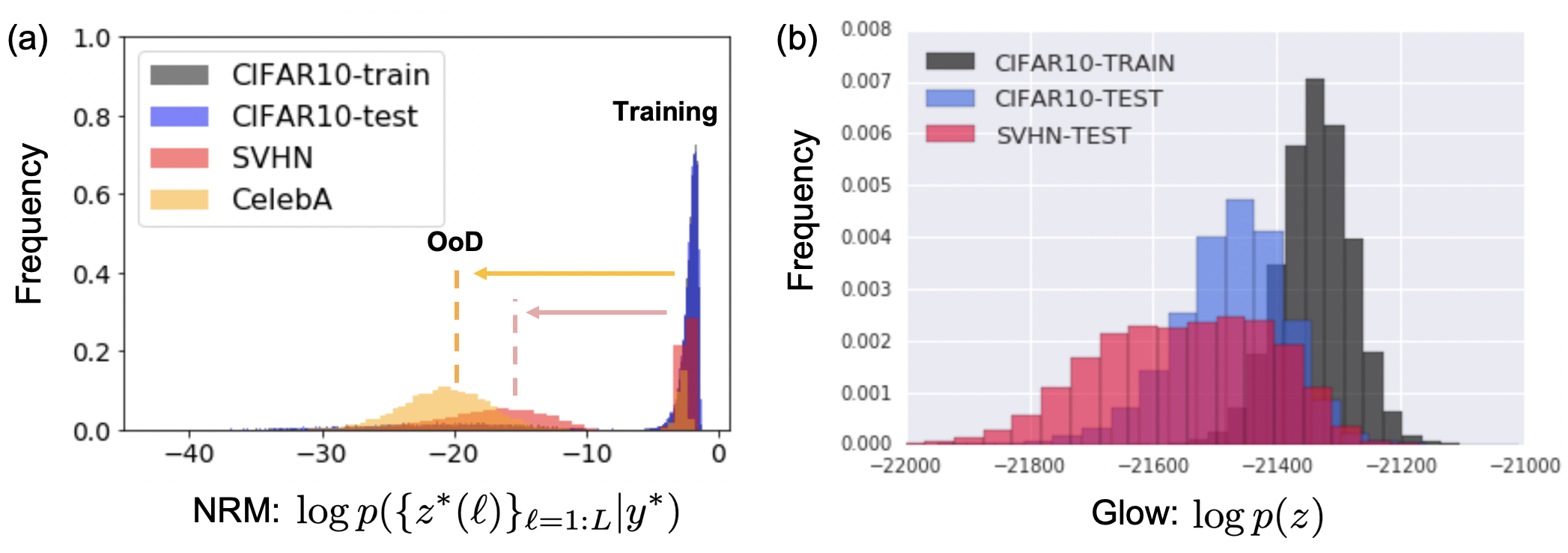}
  \caption{(a) Log joint likelihood of latent variables for NRM. Histograms for OoD data (SVHN, CelebA) lie at a lower likelihood region. In-distribution data (CIFAR10-train and CIFAR10-test) histograms overlap completely. (b) Log likelihood of latent variables for Glow. Histogram for CIFAR10-TEST has bigger overlap between OoD SVHN than CIFAR10-TRAIN, making it not a robust metric for OoD detection. }
  \label{fig:RPN}
\end{figure}
We investigate the use of data likelihood $p(x)$, reconstruction loss, and latent variable likelihood $p(z)$ as OoD detection metrics and compare their performances.

\paragraph{We propose using the joint likelihood of latent variables of the NRM as a robust metric for OoD detection.} 
The latent variables $z(\ell)$ of NRM decide whether to render a pixel and translations of rendering templates at layer $\ell$. 
The optimal values for latent variables $z^*(\ell)$ are provided by $\operatorname{ReLU}$ and $\operatorname{Maxpool}$ in the feed-forward CNN.
A structured prior $p(\{z(\ell)\}_{\ell=1:L}|y)$ is introduced to capture the dependencies of latent variables across all the layers.
Our experiment results show that $p(\{z^*(\ell)\}_{\ell=1:L}|y^*)$ is lower for OoD samples with smaller variance as shown in Figure~\ref{fig:RPN} (a), indicating that this is a robust metric for OoD detection across different datasets. 
To the best of our knowledge, this is first likelihood metric for OoD detection in deep generative models that consistently assigns lower likelihood to lower variance OoD images.
Additionally, we observe that distributions of visually similar image categories tend to overlap, suggesting that our method captures underlying structure of image data. 

\paragraph{We show that reconstruction loss at certain latent layers is able to capture the difference between distributions.} We obtain the reconstruction loss at each level of NRM and find that at some intermediate layers, it is higher for OoD data than training data.
However, the layer may vary for different datasets and model architectures, making reconstruction loss difficult to be used as a robust metric for OoD detection in practice.

\section{Approach}
We first introduce NRM by drawing a close connection between CNN. Then we show that NRM unifies likelihood and reconstruction loss based OoD detection methods by its likelihood decomposition. Finally we propose and compare the performances of three OoD detection metrics from NRM, pointing out that the joint likelihood for latent variable is a robust measure for OoD samples.
\begin{figure}[h]
  \centering
  \includegraphics[width=1.03\linewidth]{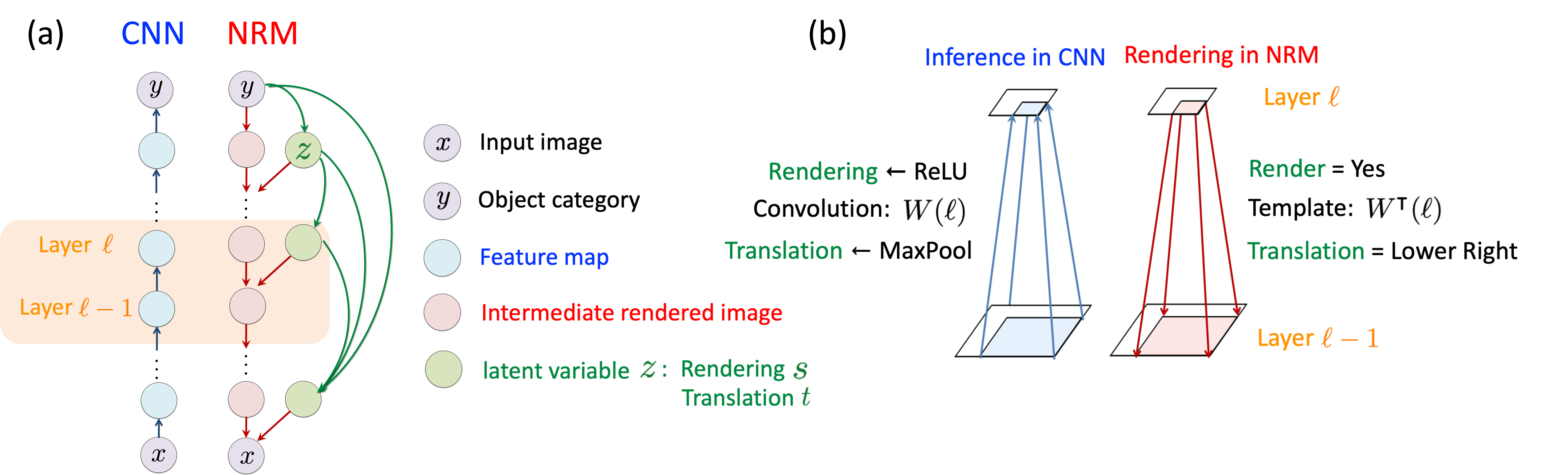}
  \caption{(a): Graphical model of NRM. Start from object category, intermediate rendered images are generated with finer details. Dependence of latent variables $z$ from layer 1 to $\ell$ are captured by a structured prior. (b) Rendering process from layer $\ell$ to layer $\ell-1$ in NRM and its connection to inference in CNN.}
  \label{fig:NRM}
\end{figure}
\subsection{Neural Rendering Model}\label{NRM-generation}
In the feed-forward process of standard CNN, information is gradually reduced to obtain the predicted label. Given the success of CNN in image classification tasks, we assume that a generative model with similar architectures to CNN is also suitable for fitting high dimensional distributions like images. NRM does so by generating images from rough to fine details using features learned from CNN as rendering templates. It introduces latent variables to model the uncertainty of rendering process and a structured prior to capture the dependencies between latent variable across layers. The graphical model of NRM is shown in Figure \ref{fig:NRM}(a). 

Let $x$ be the generated image, $y \in \{1,...,K\}$ be object category. $z(\ell)=\{t(\ell),s(\ell)\}, \ell=1,...,L$ are the latent variables at layer $\ell$, where $t(\ell)$ defines translation of rendering templates based on the position of local maximum from $\operatorname{Maxpool}$, and $s(\ell)$ decides whether to render a pixel or not based on whether it is activated ($\operatorname{ReLU}$) in the feed-foward CNN. 
$h(\ell)$ is the intermediate rendered image at layer $\ell$. $h(\ell, p), s(\ell,p)$ and $t(\ell,p)$ denotes the value of corresponding vectors at pixel $p$.

The rendering process from layer $\ell$ to layer $\ell-1$ is shown in Figure \ref{fig:NRM} (b). First, $h(\ell)$ is elementwisely multiplied by $s(\ell)$ so that only activated pixels in forward CNN are rendered in NRM. Then for each pixel, we do the following operations: 
\begin{itemize}
    \item Multiply the rendering template $W(\ell)^\intercal$ by pixel value $h(\ell, p)$, where $W(\ell)$ is the weight matrix at layer $\ell$ in CNN that contains the features learned from the data.
    \item Pad the templates with zeros to match the size of $h(\ell-1)$. Let $B(\ell)$ be the padding matrix.
    \item Translate the template to the position of local maximum from $\operatorname{Maxpool}$. We use $T(t(\ell, p))$ for the translation matrix. 
\end{itemize}
Finally, add all padded and translated templates together to get intermediate rendered image $h(\ell-1)$ at layer $\ell-1$.

Mathematically, the generation process in NRM~\cite{DBLP:journals/corr/abs-1811-02657} is as follows:
\begin{align*} 
h(\ell-1) &= \sum_{p} T(t(\ell, p)) B(\ell) W^\intercal (\ell) \left( s(\ell, p) h(\ell, p) \right)
\\ x | z, y & \sim \mathcal{N}\left(h(0), \sigma^{2} \mathbf{1}\right) 
\end{align*}
The dependencies among latent variables $\{z(\ell)\}_{1:L}$ across different layers are captured by the structured prior
$$ \pi_{z | y} \triangleq \operatorname{Softmax}\left(\frac{1}{\sigma^{2}} \sum_{\ell=1}^{L}\langle b(\ell), s(\ell) \odot h(\ell)\rangle \right)$$

where $\operatorname{Softmax}(\eta) \triangleq \frac{\exp (\eta)}{\sum_{\eta} \exp (\eta)}$, and $b(\ell)$ corresponds the bias after convolutions in CNN.

Given all the ingredients in NRM, now we formally draw the connection between NRM and CNN in Theorem~\ref{jmap}.
\begin{theorem}[MAP Inference~\cite{DBLP:journals/corr/abs-1811-02657}]
\label{jmap}
Given that intermediate rendered images $\{h(\ell)\}_{1:L}$ are nonnegative, the joint maximum a posteriori (JMAP) inference of latent variable $z$ in NRM is the feed-forward step in CNN. 
\end{theorem}
The main takeaway from this result is that given an input image, we can get its \textit{optimal} latent representations in NRM by performing the feed-forward process in CNN. 
ReLU nonlinearities indicate the optimal value of $s(\ell)$. $s(\ell)$ detects whether a feature exists in the image or not in CNN, and correspondingly determines whether or not to render a pixel in NRM. 
Maxpool operators indicate the optimal value of $t(\ell)$. $t(\ell)$ locates features in CNN and determines the position of rendering templates in NRM. 
Intuitively, the generation process with optimal latent variables in NRM is very similar to a reversed CNN. 
Similar feedback and feed-forward connections have been observed in brain~\cite{friston2018does}. 
Here, the nice connection between NRM and CNN roots from the structured joint distribution of latent variables $p(\{z(\ell)\}_{\ell=1:L}|y) = \pi_{z|y}$,  which we find to be a strong metric for OoD detection. 

\subsection{Unifying Likelihood and Reconstruction Loss Based Approaches}\label{sec:unify}
As a generative model, NRM can also be used to estimate probability distribution of input data $p(x)$. 
In this section, we first show that the log likelihood for NRM can be decomposed into two terms: reconstruction loss and joint distribution of latent variables.
Then we propose several quantities that can be extracted from NRM and discuss their usefulness in the down-streaming task: OoD detection. 

Let $x_i$ be the $i$th input data.
The rendered image at pixel level generated using mostly likely label $y_i^*$ and optimal latent variables $z_i^*$ is denoted by $h(y_i^*,z_i^*,;0)$.
The following theorem describes the likelihood decomposition for NRM. 
\begin{theorem}[Likelihood decomposition \cite{DBLP:journals/corr/abs-1811-02657}]
\label{likelihood-decomp}
The lower bound of probability density of input $x_i$ can be approximated up to constant by the following when $\sigma \rightarrow 0,$
$$\log p(x_i) \geq \mathbb{E}_{y_i, z_i}[\log p(x_i,(y_i, z_i))] \stackrel{\operatorname{asymp}}{\approx}  -\frac{1}{2\sigma^2} ||x_i-h(y_i^*,z_i^*,;0)||^2+ \log \pi_{z_i^*|y_i^*},$$
\end{theorem}
The main take away from this theorem is as follows:
\begin{itemize}
    \item The first term is proportional to reconstruction loss between the the input image and generated image using the optimal latent variables. 
    \item The second term is log joint likelihood of latent variable, which is proportional to $\sum_{\ell=1}^{L}\langle b(\ell), s(\ell) \odot h(\ell)\rangle$ up to constant.
    \item The likelihood decomposition in Theorem~\ref{likelihood-decomp} is performed at pixel level. In fact, we can also obtain latent likelihood for intermediate feature maps in CNN. 
    Let $g(x_i;k)$ be the intermediate feature map at layer $k$ of input data $x_i$ in CNN.
    The log likelihood of $g(x_i,k)$ can be lower bounded up to constant by $-\frac{1}{2\sigma^2} ||g(x_i;k)-h(y_i^*,\Tilde{z_i},;k)||^2+ \log \pi_{\Tilde{z_i}|y_i^*},$
    where $\Tilde{z_i}=\{z_i^*(\ell)\}_{\ell=k:L}.$
\end{itemize}

As stated in Theorem~\ref{likelihood-decomp}, the NRM provides estimations of three quantities related to input data: data likelihood, reconstruction loss and joint likelihood of latent variables.
We discuss how they can be used for OoD detection.
\\\textbf{Log likelihood of data $\log p(x)$ }
As the direct modeling of training data distribution, likelihood seems to be the most straightforward metric for OoD detection. However, previous works have shown that higher likelihood could be assigned to OoD samples. We analyze why this is the case in our experiments (Section \ref{nrmlikelihood}). 
\\\textbf{Reconstruction loss at all layers }
Since NRM closely resembles the architecture of CNN, we can obtain the reconstruction loss for feature maps at each layer in addition to the input image layer. 
Since CNNs have rich representations at intermediate layers \cite{DBLP:journals/corr/ZeilerF13}, reconstruction loss at those layers may be a useful OoD metric. We analyze this metric in Section \ref{sec:reconst}.
\\\textbf{Log joint likelihood of latent variables $p(\{z^*(\ell)\}_{\ell=1:L}|y^*)$ } 
This is our proposed method for OoD detection. From the forward process of CNN, we can find the most likely label $y^*$ and the most likely set of latent variables $\{z^*\}_{\ell=1:L}$ to compute this value.
Intuitively, the set of latent variables $\{z^*_i(\ell)\}_{\ell=1:L}$ defines a "rendering path" for an input image. 
The "rendering path" specifies which pixels to render and the locations for rendering templates.  We observe the performance of this metric in Section \ref{sec:rpn}.

\section{Experiments}
\subsection{Experimental Set-Up}
NRM can be used for diverse networks within the ConvNet family. In our experiments, we use the NRM architecture whose inference corresponding to the All Convolutional Net introduced in  \cite{springenberg2014striving}.
We preprocess our data by scaling the images into the range [-1, 1].  We use CIFAR-10 \cite{krizhevsky2009learning} as the in-distribution dataset.  For OoD datasets, we use SVHN \cite{netzer2011reading}, CIFAR-100 \cite{krizhevsky2009learning}, and CelebA ~\cite{liu2015faceattributes}.  For the CelebA dataset, we use the first 10k aligned and cropped images and multiply the data by $0.8$ so that the variance is smaller than that of CIFAR-10.
We perform experiments using the three OoD detection metrics described in Section \ref{sec:unify} on different datasets. A summary of these metrics is shown in Table \ref{summary-table}.
\begin{table}[ht]
  \caption{Summary of mean value of OoD detection metrics over 3 runs.}
  \label{summary-table}
  \centering
  \begin{tabular}{cccc}
    \toprule
                    & $\log p(x)$    &  $\log p(\{z^*(\ell)\}_{\ell=1:L}|y^*)$ & \begin{tabular}{@{}c@{}}Negative reconstruction \\ loss at layer 6 \footnotemark \end{tabular} \\
    \midrule
    CIFAR10-train   & $\mathbf{-55.47 \pm 7.78}$         & $-5.73 \pm 0.25$       & $\mathbf{-5.839}$\\
    CIFAR10-test    & $-55.63 \pm 7.89$         & $\mathbf{-5.66 \pm 0.25}$        & $-5.846$\\
    SVHN            & $-59.64 \pm 10.16$         & $-8.16 \pm 1.45$        & $-10.888$\\
    CelebA          & $-58.69 \pm 7.82$        & $-17.60 \pm 0.92$        & $-8.008$\\
    \bottomrule
  \end{tabular}
\end{table}
\footnotetext{Standard deviation calculations are not included since layer at which separation can be seen varies between models}

\subsection{Problems with Likelihood}\label{nrmlikelihood}
\begin{figure}[ht]
  \centering
  \includegraphics[width=1.1\linewidth]{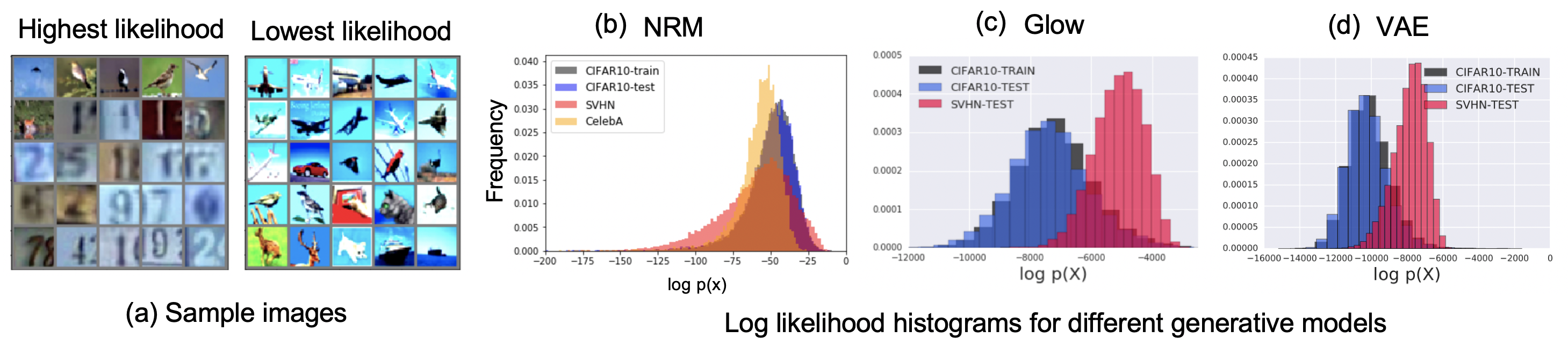}
  \caption{(a) Top 25 images with highest and lowest likelihood from NRM. Log-likelihood histograms for (b) NRM, (c) Glow and (d) VAE trained on CIFAR-10 and tested on SVHN. (histogram for Glow and VAE from \cite{nalisnick2018deep}). We also use CelebA as OoD data for NRM. While Glow and VAE assign higher likelihood for OoD samples from SVHN, the likelihood for OoD samples is similar to that of training data using NRM.}
  \label{fig:top-25}
\end{figure}
We first show that likelihood $p(x)$ is not a strong metric for OoD detection since it does not capture image structure. We found that in practice $p(x)$ is dominated by the reconstruction loss term in the likelihood decomposition.  From Table \ref{summary-table}, we can see that the mean of $\log p(x)$ is lower for SVHN than CIFAR-10. However, observing the spread of $\log p(x)$ in Figure \ref{fig:top-25}. We can see that the histograms for SVHN and CIFAR-10 share a large overlap, and the peak of SVHN lies within the peak of CIFAR-10 despite the fact that SVHN is OoD. In order to understand why this occurs, we visualized the top 25 images with the highest and lowest likelihood for an NRM trained on CIFAR-10 in Figure \ref{fig:top-25} (a).  We notice that the brightness of the background contributes a lot to the likelihood. Images with duller background tend to have higher likelihood than those with high contrast. Intuitively, it is because images with low variance and pixel values closer to the mean image can be easily reconstructed, resulting in a lower reconstruction term in the likelihood decomposition.  Therefore, likelihood estimation based on pixel space distributions is not reliable for OoD detection as it tends to distinguish data points based on their low dimensional statistics (mean, variance of pixels) rather than their semantic meaning.

\subsection{Joint Distribution of Latent Variables}\label{sec:rpn}
We show that our proposed metric, the joint distribution of latent variables $p(\{z^*(\ell)\}_{\ell=1:L}|y^*)$, is a reliable metric for OoD detection. $y^*$ is the most likely label and $\{z^*(\ell)\}_{\ell=1:L}$ is the most likely set of latent variables.  These latent variables are obtained through the feed-forward CNN by Theorem \ref{jmap}.  Since this theorem assumes non-negativity of intermediate rendered images $h(\ell)$, we use a modified training process to reduce negativity in $h(\ell)$.  More details about the training process are described in Appendix \ref{appendix: training}.
\paragraph{Performance on Dissimilar Datasets}
To show the consistency of our method, we visualize the performance by plotting histograms of $\log p(\{z^*(\ell)\}_{\ell=1:L}|y^*)$ (shown in Figure~\ref{fig:RPN}) for in and out of distribution data.  
Unlike the likelihood histograms in figure \ref{fig:top-25}, we can see separation in peaks of the distributions of $\log p(\{z^*(\ell)\}_{\ell=1:L}|y^*)$. We note that the mean of $\log p(\{z^*(\ell)\}_{\ell=1:L}|y^*)$ is consistently lower for these OoD datasets while the histograms for in-distribution data, CIFAR-10 train and test sets, lie on top of each other.

Unlike the likelihood histograms in Figure \ref{fig:top-25}, the peaks of OoD histograms lie to the left of the peaks of the in-distribution histograms. This is intuitive because $p(\{z^*(\ell)\}_{\ell=1:L}|y^*)$ is the likelihood of using a specific combination of latent variables when rendering from a label. Even if the model is able to reconstruct an OoD image well, it is unlikely that the chosen combination of latent variables will occur naturally from the original distribution, making this method is less sensitive to image variance.   We can see that this feature is consistent across OoD distributions of smaller variance (SVHN, CelebA).  To the best of our knowledge, this is the first time this has been achieved using a likelihood method.

We contrast the distribution of $\log p(\{z^*(\ell)\}_{\ell=1:L}|y^*)$ to the distribution of $p(z)$ observed in Glow by \cite{nalisnick2018deep} as shown in Figure \ref{fig:RPN}. The $z$ in the $p(z)$ of Glow corresponds to the last layer of latent variables. In Glow's $p(z)$ distribution, we observe some separation between the peaks of in-distribution images (CIFAR-10) and OoD images (SVHN) but also observe similar amount of separation between the peaks of train and test set of the in-distribution images. This separation suggests that the final layer of latent variables captures details which are specific to the training set instead of general features of the whole data distribution. In contrast, the joint distribution $\log p(\{z^*(\ell)\}_{\ell=1:L}|y^*)$ does not suffer from this problem because it uses information from the latent variables across all layers, causing it to not be overly sensitive to features only in the training set. Thus, $\log p(\{z^*(\ell)\}_{\ell=1:L}|y^*)$ is a better distinguisher between in-distribution and OoD samples.

\begin{figure}[ht]
  \centering
  \includegraphics[width=1\linewidth]{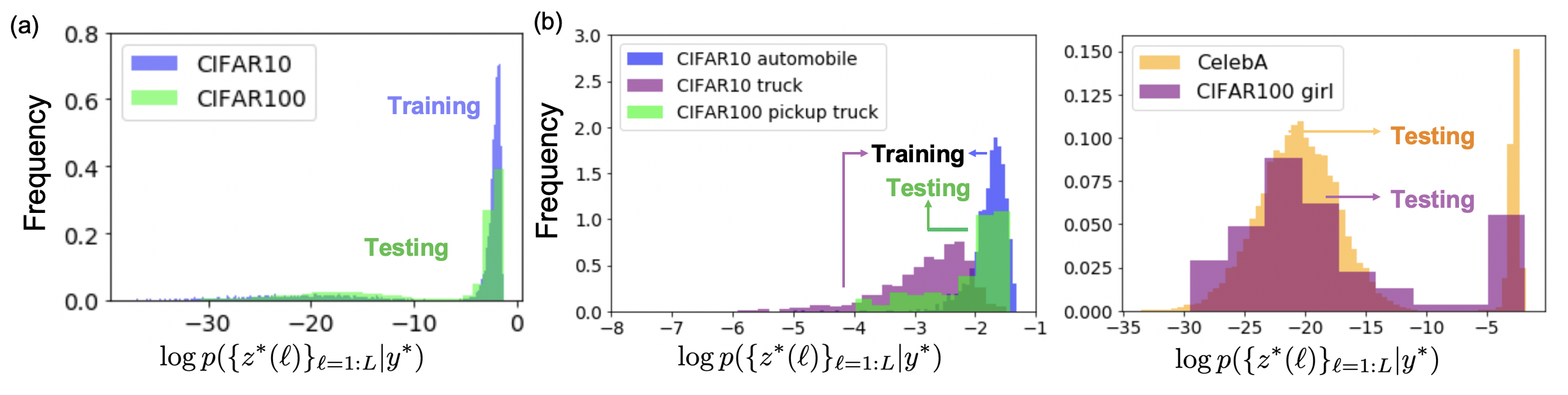}
  \caption{(a) Histograms of $p(\{z^*(\ell)\}_{\ell=1:L}|y^*)$ of CIFAR-10 and CIFAR-100 using NRM trained on CIFAR-10. We observe that CIFAR-10 shares a large overlap with CIFAR-100. (b) Left: A histogram of $p(\{z^*(\ell)\}_{\ell=1:L}|y^*)$ of CIFAR-10 automobile, CIFAR-10 truck, and CIFAR-100 pickup truck. Right: histogram of $p(\{z^*(\ell)\}_{\ell=1:L}|y^*)$ for the girl category of CIFAR-100 and CelebA.  We see that the distribution of $p(\{z^*(\ell)\}_{\ell=1:L}|y^*)$ aligns for these visually similar categories, indicating that image structure is captured.}
  \label{fig:overlap}
\end{figure}

\paragraph{Structure-based Detection}
Our method captures image structure within the latent variables, we plot histograms of $\log p(\{z^*(\ell)\}_{\ell=1:L}|y^*)$ for similar datasets. Since CIFAR-10 and CIFAR-100 are both subsets of the Tiny Images dataset, we expect these datasets to be more similar to each other than to SVHN or CelebA.  From Figure~\ref{fig:overlap} (a), we can see that CIFAR-10 and CIFAR-100 share a much larger area of overlap than SVHN or CelebA in Figure \ref{fig:RPN}. The area of overlap is indicative of dataset similarity.  To further analyze this feature of our metric, we generate category-specific histograms.  A histogram of the CIFAR-10 automobile, CIFAR-10 truck, and CIFAR-100 pickup truck categories are portrayed in ~\ref{fig:overlap} (b). We can see that although the pickup truck class of CIFAR-100 is OoD, the histogram of $\log p(\{z^*(\ell)\}_{\ell=1:L}|y^*)$ overlaps with the automobile and truck classes of CIFAR-10. We also observe large overlap between CelebA and the girl category of CIFAR-100.  These large overlaps are indicative of image similarity, suggesting that our metric is able to capture the structure of images well.  This contrasts the traditional likelihood in which we saw that background color rather than the image content influences the likelihood estimate.

\subsection{An Evaluation of Reconstruction Loss}\label{sec:reconst}
\begin{figure}[ht]
  \centering
  \includegraphics[width=1.0\linewidth]{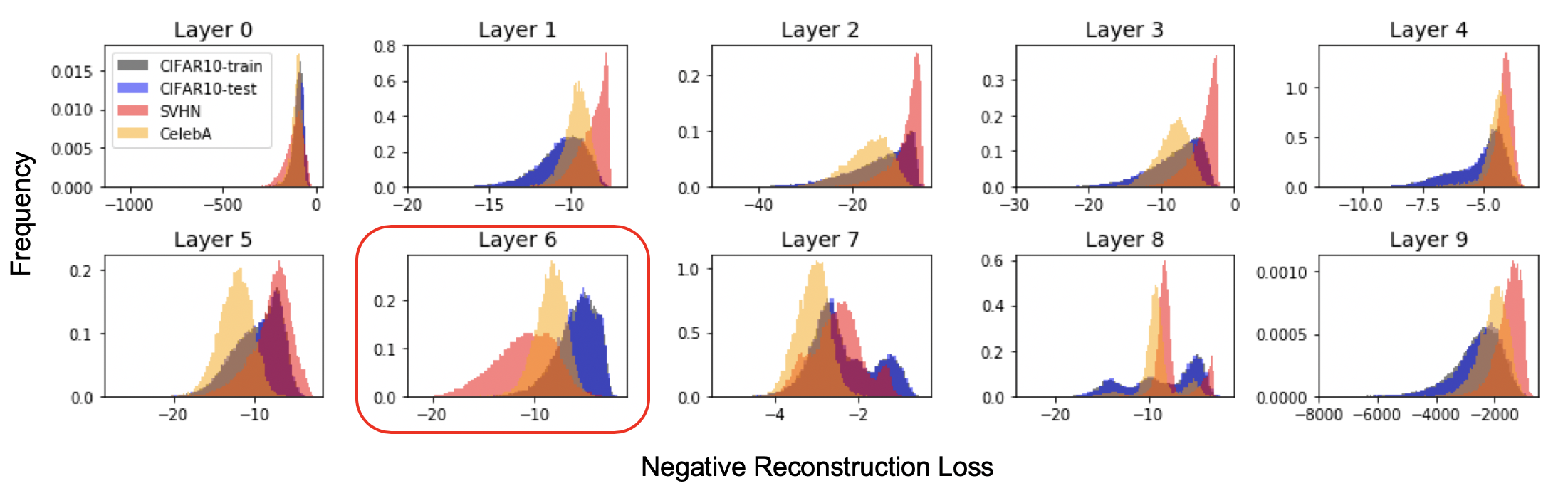}
  \caption{Reconstruction loss at different layers. Layer 0 to Layer 9: pixel level to one layer above the labels. At layer 6, the reconstruction loss is higher for SVHN and CelebA, which could be useful for OoD detection. But in practice it is hard to choose which layer to look at before seeing OoD data because the layer may vary for different network architectures and datasets.}
  \label{fig:recons}
\end{figure}
To demonstrate the performance of reconstruction loss at different layers of the NRM, we plot histograms of this metric at each layer, shown in Figure~\ref{fig:recons}.  We note that at pixel level, the histograms for all the datasets overlap, suggesting that the model is complex enough to reconstruct any input image. We observe that at some intermediate layers, for example layer 6, OoD distributions (SVHN, CelebA) have larger reconstruction loss than CIFAR-10. 
This is because the intermediate layers in CNN have richer representations of data, it is more difficult for the model to reconstruct those layers than the extreme layers. 
However, we usually do not know about the OoD distribution or which layers are sensitive to the difference between in-distribution samples and the OoD samples, this poses difficulties in using reconstruction loss in practice.  In comparison, our proposed method of using $p(\{z^*(\ell)\}_{\ell=1:L}|y^*)$ does not need any prior knowledge of the OoD distribution, and consistently shows a meaningful separation between peaks of the histograms for in-distribution and OoD images.

\section{Discussion}
A major strength of our method lies in the fact that it captures the structure within images, causing it to be less influenced by image variance. Overlaps within the distributions of $p(\{z^*(\ell)\}_{\ell=1:L}|y^*)$ are also interpretable: large overlaps represent high similarity between datasets.  In addition, $p(\{z^*(\ell)\}_{\ell=1:L}|y^*)$ has an intuitive explanation: we can think of it as uncertainty in the choice of latent variables during the rendering process. We will now compare our method to other existing metrics for deep networks and propose future directions for research.
\paragraph{Comparison with Other OoD Metrics for Deep Networks}
A recent study by \cite{DBLP:journals/corr/abs-1812-04606} shows that OoD detection using deep networks can be improved by feeding the model with OoD samples.  However, in practice, the OoD distribution is often unknown.  Our method works under the challenging setting where we do not have prior knowledge on the OoD distribution.  We hypothesize that our model will perform better if we do have access to OoD samples during training.

Another study by \cite{DBLP:journals/corr/abs-1902-02767} proposes neural hybrid model consisting of a linear model defined on a set of features by a flow-based model. Their findings show a complete separation between the distribution of $\text{log }p(x)$ between SVHN and CIFAR-10 distributions when trained on SVHN.  Although our results do not show a clear-cut a separation, we focus on the more difficult direction: assigning lower likelihood to SVHN (OoD samples with smaller variance) when trained on CIFAR-10.  In addition, our method for OoD detection works for CNNs with ConvNet structure such as AlexNet, ResNet, and the All Convolutional Net, which are used more in practice than flow-based generative models.

\cite{choi2018generative} explores the use of Watanabe Akaike Information Criterion (WAIC) for OoD detection.  Unlike our method which is a pure likelihood metric, WAIC is a hybrid metric calculated in terms of likelihood. This metric measures the gap between training and test distributions and has been shown to be effective for OoD detection with ensembles of GANs. Additionally it is shown to assign lower score to OoD image distributions with lower variance.  Similar to our method, there is also overlap between distributions.
\paragraph{Future Directions}

As can be seen from Figure \ref{fig:RPN}, the histograms of $\text{log } p(\{z^*(\ell)\}_{\ell=1:L}|y^*)$ still have overlap. The size of these overlaps should be reduced for robust OoD detection. A possible solution is weighting this metric during training. If we weight this term more highly, the model will try to reduce the uncertainty in the latent variables due to higher penalty, leading to more in-distribution class specific latent variables.

Another possible direction is using a hybrid metric in terms of latent variable likelihood to achieve better separation. For example, we can calculate WAIC in terms of our metric. Since our metric improves on traditional likelihood used by WAIC, this may lead to better separation between OoD samples and training samples.

Since we focused our research mainly on the difficult situation where the OoD data has less variance, it would be also meaningful to see how our metric performs in easier situations where the image variance is similar or higher. For instance, comparing our metric to state-of-the-art methods for OoD detection on different categories of MNIST.

\section{Conclusion}
We show that using neural rendering methods for OoD detection unifies likelihood and reconstruction based approaches to OoD detection. We show that while likelihood and reconstruction are difficult to use for OoD detection due to the sensitivity to pixels rather than to image structure, the joint likelihood of latent variables $p(\{z^*(\ell)\}_{\ell=1:L}|y^*)$ can capture image structure better and be used as a metric for OoD data. We find that this metric correctly assigns lower average likelihood to OoD images with lower variance than in-distribution images and capture similarities between distributions within overlaps. To the best of our knowledge, our metric is first to do so for deep generative models.

\subsubsection*{Acknowledgments}
A. Anandkumar is supported in part by Bren endowed chair, Darpa PAI, Raytheon, and Microsoft, Google and Adobe faculty fellowships.

\medskip

\small

\bibliographystyle{apalike}
\bibliography{refs}

\newpage
\begin{thebibliography}{}

\bibitem[An and Cho, 2015]{an2015variational}
An, J. and Cho, S. (2015).
\newblock Variational autoencoder based anomaly detection using reconstruction
  probability.
\newblock {\em Special Lecture on IE}, 2:1--18.

\bibitem[Choi and Jang, 2018]{choi2018generative}
Choi, H. and Jang, E. (2018).
\newblock Generative ensembles for robust anomaly detection.
\newblock {\em arXiv preprint arXiv:1810.01392}.

\bibitem[Friston, 2018]{friston2018does}
Friston, K. (2018).
\newblock Does predictive coding have a future?
\newblock {\em Nature neuroscience}, 21(8):1019.

\bibitem[Guo et~al., 2017]{guo2017calibration}
Guo, C., Pleiss, G., Sun, Y., and Weinberger, K.~Q. (2017).
\newblock On calibration of modern neural networks.
\newblock In {\em Proceedings of the 34th International Conference on Machine
  Learning-Volume 70}, pages 1321--1330. JMLR. org.

\bibitem[Hendrycks et~al., 2018]{DBLP:journals/corr/abs-1812-04606}
Hendrycks, D., Mazeika, M., and Dietterich, T.~G. (2018).
\newblock Deep anomaly detection with outlier exposure.
\newblock {\em CoRR}, abs/1812.04606.

\bibitem[Ho et~al., 2018]{DBLP:journals/corr/abs-1811-02657}
Ho, N., Nguyen, T., Patel, A.~B., Anandkumar, A., Jordan, M.~I., and Baraniuk,
  R.~G. (2018).
\newblock Neural rendering model: Joint generation and prediction for
  semi-supervised learning.
\newblock {\em CoRR}, abs/1811.02657.

\bibitem[Kingma and Dhariwal, 2018]{kingma2018glow}
Kingma, D.~P. and Dhariwal, P. (2018).
\newblock Glow: Generative flow with invertible 1x1 convolutions.
\newblock In {\em Advances in Neural Information Processing Systems}, pages
  10215--10224.

\bibitem[Krizhevsky and Hinton, 2009]{krizhevsky2009learning}
Krizhevsky, A. and Hinton, G. (2009).
\newblock Learning multiple layers of features from tiny images.
\newblock Technical report, Citeseer.

\bibitem[Li et~al., 2018]{li2018anomaly}
Li, D., Chen, D., Goh, J., and Ng, S.-k. (2018).
\newblock Anomaly detection with generative adversarial networks for
  multivariate time series.
\newblock {\em arXiv preprint arXiv:1809.04758}.

\bibitem[Liu et~al., 2015]{liu2015faceattributes}
Liu, Z., Luo, P., Wang, X., and Tang, X. (2015).
\newblock Deep learning face attributes in the wild.
\newblock In {\em Proceedings of International Conference on Computer Vision
  (ICCV)}.

\bibitem[Nalisnick et~al., 2018]{nalisnick2018deep}
Nalisnick, E., Matsukawa, A., Teh, Y.~W., Gorur, D., and Lakshminarayanan, B.
  (2018).
\newblock Do deep generative models know what they don't know?
\newblock {\em arXiv preprint arXiv:1810.09136}.

\bibitem[Nalisnick et~al., 2019]{DBLP:journals/corr/abs-1902-02767}
Nalisnick, E.~T., Matsukawa, A., Teh, Y.~W., G{\"{o}}r{\"{u}}r, D., and
  Lakshminarayanan, B. (2019).
\newblock Hybrid models with deep and invertible features.
\newblock {\em CoRR}, abs/1902.02767.

\bibitem[Netzer et~al., 2011]{netzer2011reading}
Netzer, Y., Wang, T., Coates, A., Bissacco, A., Wu, B., and Ng, A.~Y. (2011).
\newblock Reading digits in natural images with unsupervised feature learning.

\bibitem[Springenberg et~al., 2014]{springenberg2014striving}
Springenberg, J.~T., Dosovitskiy, A., Brox, T., and Riedmiller, M. (2014).
\newblock Striving for simplicity: The all convolutional net.
\newblock {\em arXiv preprint arXiv:1412.6806}.

\bibitem[Zeiler and Fergus, 2013]{DBLP:journals/corr/ZeilerF13}
Zeiler, M.~D. and Fergus, R. (2013).
\newblock Visualizing and understanding convolutional networks.
\newblock {\em CoRR}, abs/1311.2901.

\bibitem[Zenati et~al., 2018]{zenati2018efficient}
Zenati, H., Foo, C.~S., Lecouat, B., Manek, G., and Chandrasekhar, V.~R.
  (2018).
\newblock Efficient gan-based anomaly detection.
\newblock {\em arXiv preprint arXiv:1802.06222}.

\bibitem[Zhou and Paffenroth, 2017]{zhou2017anomaly}
Zhou, C. and Paffenroth, R.~C. (2017).
\newblock Anomaly detection with robust deep autoencoders.
\newblock In {\em Proceedings of the 23rd ACM SIGKDD International Conference
  on Knowledge Discovery and Data Mining}, pages 665--674. ACM.

\end{thebibliography}

\newpage
\appendix
\appendixpage
\addappheadtotoc

\renewcommand\thefigure{\thesection.\arabic{figure}}    
\setcounter{figure}{0}

\section{A closer look at NRM}
The notations used to define NRM are summarized in Table \ref{notation-NRM}. The generation process in NRM is described in Section \ref{NRM-generation}. Given a class category $y$, NRM samples latent variables from a structured prior $\pi_{z|y}$.  Then, NRM starts to render images from coarse to fine details by a sequence of linear transformations defined by $\{s(\ell), t(\ell)\}, W^\intercal(\ell)$ and $B(\ell)$. The finest image $h(y,z;0)$ is rendered at the bottom of NRM. Finnally, Gaussian pixel nosie is added to render the final image $x$ as $x | z, y \sim \mathcal{N}\left(h(y,z; 0), \sigma^{2} \mathbf{1}\right) $.

Although the above generation process is complex enough to represent any input image, it is hard to generate natural images in reasonable time because of the huge number degrees of freedom of latent variables. Therefore, it is necessary to incorporate the prior knowledge we have on natural images to better structure the NRM. Classification models like CNN have been known to be able to get good representations for natural images through feature maps. Thus, NRM is designed so that its inference for optimal latent variables $z^*$ yields the feed-forward process in CNN. Similarly, $y^*$ used to reconstruct an unlabeled image is the mostly likely label from CNN. 
\begin{table}[h]
  \caption{Summary of notations in NRM.}
  \label{notation-NRM}
  \centering
  \begin{tabular}{lll}
    \toprule
    Notations    &Variables      &  Size  \\
    \midrule
    $x$         & input image   & $D(0)$ \\
    $y$         & object category  & 1            \\
    $z(\ell)$   & all latent variables in layer $\ell$, i.e. $\{s(\ell), t(\ell)\}$ & $\{D(\ell), D(\ell)\}$ \\
    $s(\ell)$   & rendering latent variables in layer $\ell$  & $D(\ell)$  \\
    $s(\ell,p)$ & rendering latent variables in layer $\ell$ at pixel location $p$  & 1  \\
    $t(\ell)$   & translation latent variables in layer $\ell$  & $D(\ell)$  \\
    $t(\ell,p)$ & translation latent variables in layer $\ell$ at pixel location $p$  & 1  \\
    $h(\ell)$   & intermediate rendered image in layer $\ell$ in general & $D(\ell)$ \\
    $h(y,z; \ell)$   & \begin{tabular}{@{}l@{}}intermediate rendered image in layer $\ell$ from object \\ category $y$ using latent variables $z$ \end{tabular}
    & $D(\ell)$ \\
    $y^*$       & most likely label for unlabeled data & 1 \\
    $z^*(\ell)$ & optimal latent variables from feed-forward CNN & $\{D(\ell), D(\ell)\}$\\
    \toprule
    Notations    & Parameters    &  Size  \\
    \midrule
    $W(\ell)$  & weight matrix of features learned from CNN at layer $\ell$ & $1\times F(\ell)$ \\
    $B(\ell)$  & zero-padding matrix at layer $\ell$  & $D(\ell-1) \times F(\ell)$ \\
    $T(t(\ell,p))$ & \begin{tabular}{@{}l@{}}translation matrix chosen according to the value of\\
     translation latent variable $t(\ell,p)$ \end{tabular} & $D(\ell-1) \times D(\ell-1)$  \\
    $b(\ell)$  &  bias term after convolutions in CNN at layer $\ell$ & $D(\ell)$\\
    $\sigma^2$ & pixel noise variance & 1\\
    \bottomrule
  \end{tabular}
\end{table}

\section{Training procedure} \label{appendix: training}
For Theorem \ref{jmap} to hold, the intermediate rendered images $\{h(\ell)\}_{\ell=1:L}$ need to be non-negative. Under the non-negativity assumption, the inferred latent variables from feed-forward CNN are exact. However, this assumption does not necessarily hold in practice. In order to approximate the optimal latent variables, we can minimize the negativity loss $\sum_{\ell=1}^L||h_-(\ell)||^2$ during training the model, where $h_-(\ell)$ is the negative part of intermediate rendered image $h(\ell)$. We find that this step is crucial for OoD detection using $p(z^*|y^*)$. 

\section{A visualization of intermediate layers}
\begin{figure}[h]
  \centering
  \includegraphics[width=0.75\linewidth]{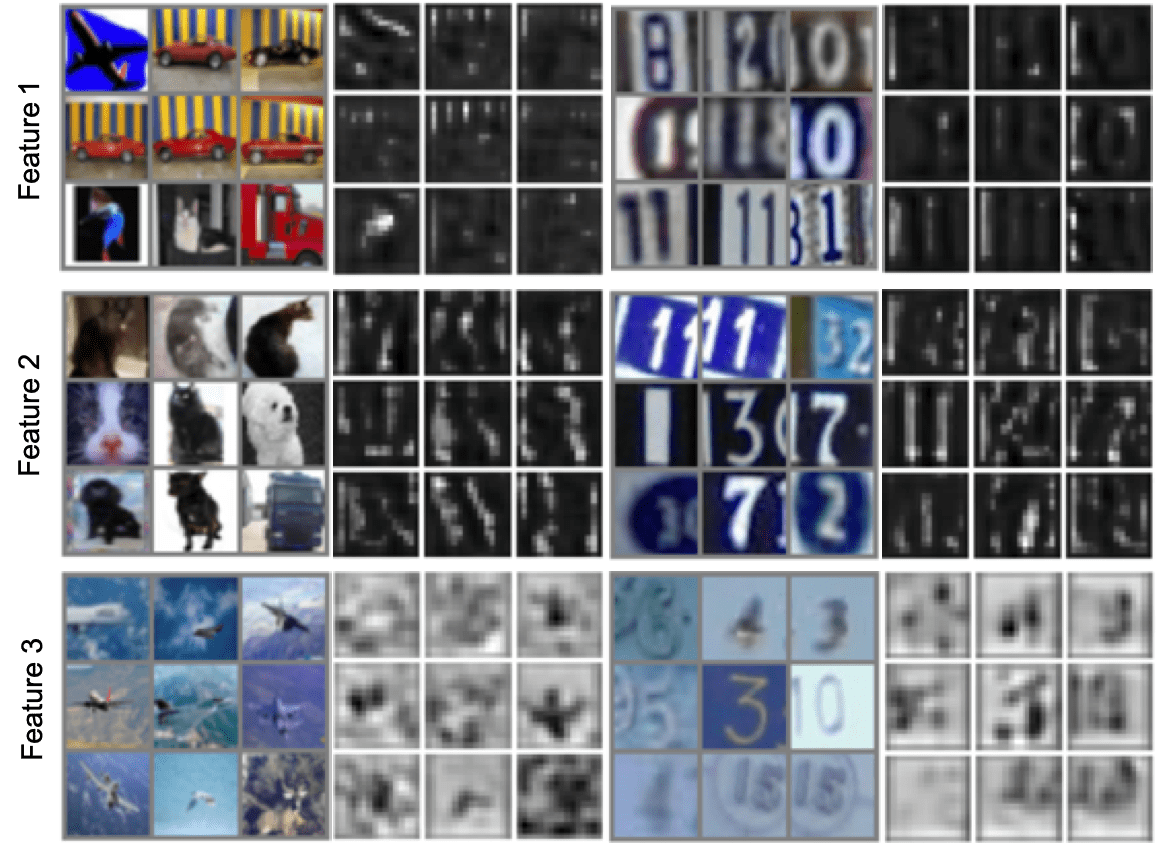}
  \caption{Visualization of top 9 in-distribution images (CIFAR-10) and OoD images (SVHN) with highest activation values for randomly selected features at intermediate layer 6.}
  \label{fig:features}
\end{figure}

Figure \ref{fig:features} shows images with highest activation values for a randomly selected subset of features at layer 6 (The layer that consistently shows higher reconstruction loss for OoD samples). We see grouping within each feature and different features focus on different patterns. For instance, feature 1 focuses on striped structures, feature 2 detects the edges between foreground objects and background, and feature 3 is related to blueish backgrounds. In CNN, the first several layers tend to focus on large patch structures and the last several layers would exaggerate on details. The intermediate layer contains the richer representations of input data and thus it is more difficult to reconstruct OoD samples at intermediate layers.

\section{NRM's reconstruction ability}

\paragraph{Reconstruction for high likelihood and low likelihood images}

\begin{figure}[ht]
  \centering
  \includegraphics[width=1.0\linewidth]{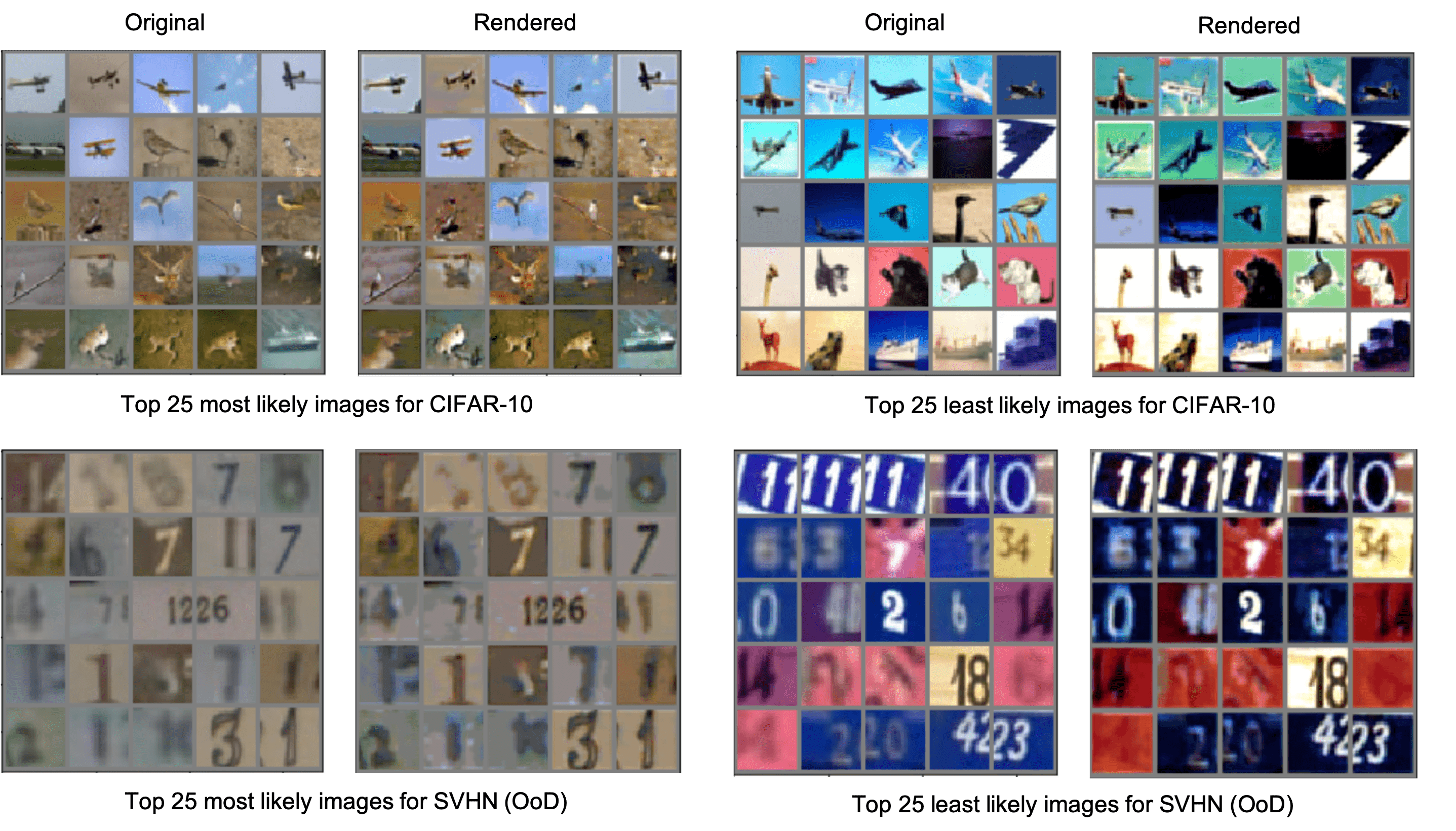}
  \caption{Original and rendered images from NRM for most and least likely in-distribution samples (CIFAR-10) and OoD samples (SVHN).}
  \label{fig:rendered}
\end{figure}

We visualize the relationship between likelihood and reconstruction loss by generating the rendered images for the top 25 most likely and least likely images shown in Figure \ref{fig:rendered}. As we showed before, background color influences likelihood.  We can see that when the background color is duller, the rendered images look almost exactly like the original image. When the background is brighter, the model is unable to render the same color pixels.  Here, we can see that higher reconstruction loss leads to lower likelihood.

\begin{figure}[ht]
  \centering
  \includegraphics[width=0.6\linewidth]{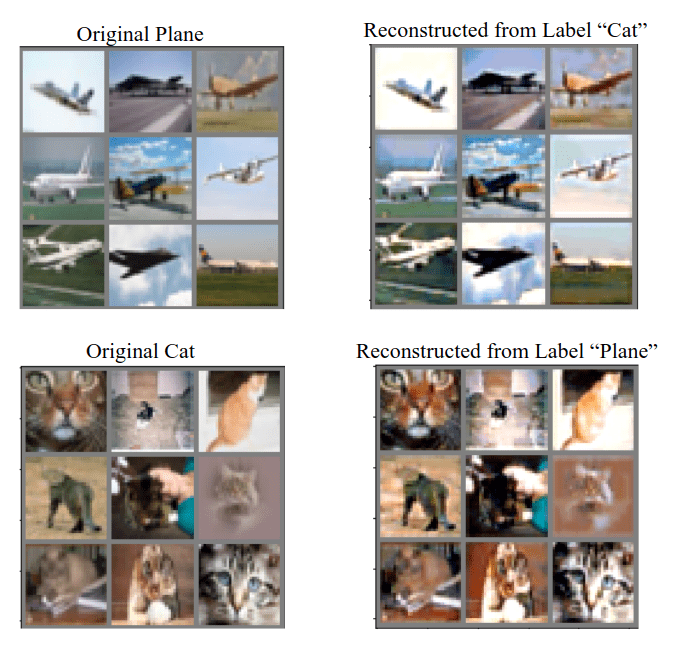}
  \caption{Top row: Original airplane image (left) vs airplane image reconstructed from label "Cat" using optimal latents for reconstructing original plane (right). Bottom row: Original cat image (left) vs cat image reconstructed from label "Airplane" using optimal latents for reconstructing original cat}
  \label{fig:falselabel}
\end{figure}

\paragraph{Reconstruction with the wrong label}
As described in Section 2.1, the NRM reconstructs images using the labels $y$ and latent variables $z(\ell) = \{t(\ell), s(\ell)\}$ corresponding to translation and rendering switches to revert $\operatorname{Maxpool}$ and $\operatorname{ReLU}$ respectively in the forward process.  We reconstruct an image $x$ using the most likely values of $t(\ell), s(\ell)$ which we obtain by passing $x$ through the forward process and record the positions of Maxpool and ReLU states. To understand how much the label $y$ contributes to the quality of the reconstructed image, we reconstruct with a false label using the optimal latent variables learned in the forward process. Two examples are shown in Figure \ref{fig:falselabel}: reconstruction of plane from false label "cat" and reconstruction of cat from false label "plane".  We see that the original and reconstructed images still have the same structure, suggesting that the information needed for good reconstruction is completely captured by the latent variables.

\section{Mean of latent variables}
\begin{figure}[ht]
  \centering
  \includegraphics[width=1.0\linewidth]{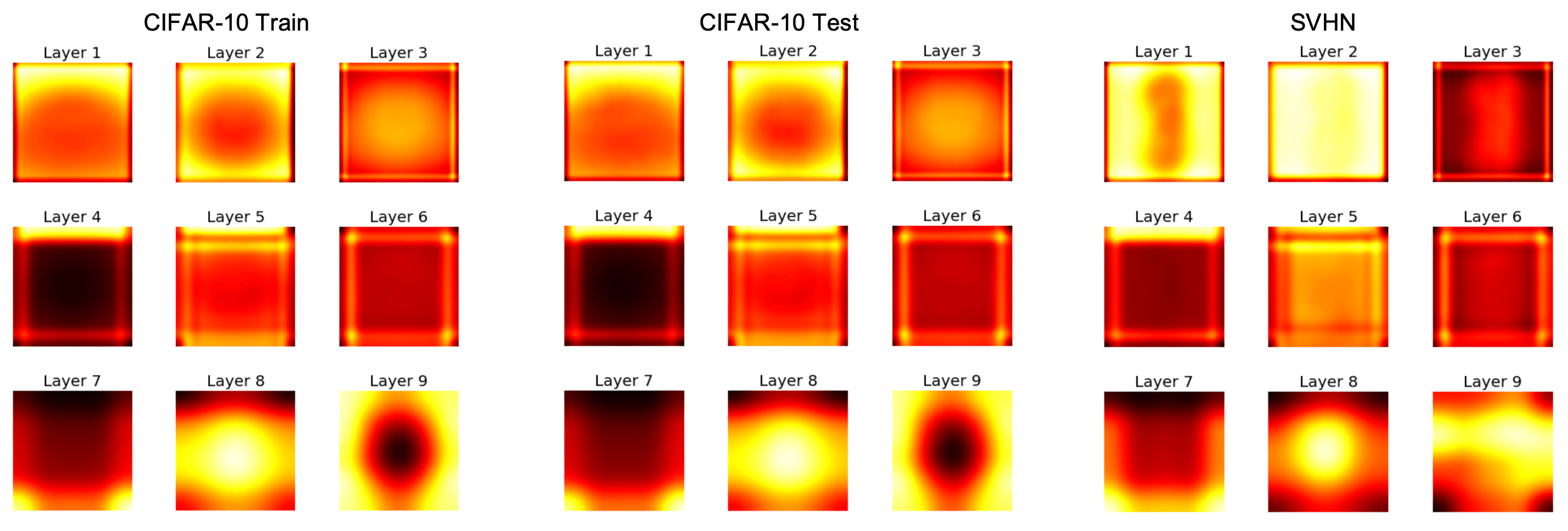}
  \caption{Mean of rendering latent variable $s(\ell)$ at each layer. We see that for CIFAR-10 train and test sets, the mean latents are almost the same. The mean latents for SVHN show conspicuous difference from those of CIFAR-10.}
  \label{fig:avg-latents}
\end{figure}

We visualize the mean of redering latent variables $s(\ell)$ at all layers for the in-distribution (CIFAR-10) and OoD distribution (SVHN) datasets in Figure \ref{fig:avg-latents}.  We see that for the CIFAR-10 train and test datasets, the mean of rendering latent variables are almost identical.  However, this is not the case for SVHN. While the NRM can reconstruct SVHN well, the rendering path (specified by latent variables) differ greatly from that for CIFAR-10 images. This shows the potential of latent variable in OoD detection.

\end{document}